\documentclass{myThesisTemp}

\usepackage{threeparttable}

\usepackage[outdir=./]{epstopdf}
\usepackage{rotating}
\usepackage{booktabs}

\title{Uncertainty-Resilient Multimodal Learning via Consistency-Guided Cross-Modal Transfer}{일관성 기반 교차모달 전이를 통한 불확실성에 대응하는 다중모달 학습}

\author[korean]{장 효 정}{}{}
\author[english]{Jang}{Hyo-Jeong}{}

\advisor{이 성 환}{Seong-Whan Lee}

\department{
Brain and Cognitive Engineering
}{
뇌 공 학 과
}

\committee[1]{Seong-Whan Lee} 
\committee[2]{Won-Zoo Chung}
\committee[3]{Tae-Eui Kam}

\graduateDate{2026}{2}
\submitDate{2025}{10}
\approvalDate{2025}{12}

\captionLineSpacing{150}
\abstractLineSpacing{200}
\krAbstractLineSpacing{200}
\TOCLineSpacing{200}
\contentLineSpacing{200}
\acknowledgementLineSpacing{200}

\begin{document}


\chapter{Introduction}

\sloppy
Multimodal learning has become increasingly important in brain–computer interfaces (BCI) and human–computer interaction (HCI) systems, where models must interpret a wide range of human behaviors and internal states expressed through diverse sensory channels~\cite{review_paradigms, human2006}. Human interactions with such systems generate multimodal signals—such as physiological measurements, visual cues, audio expressions, and contextual information—each capturing complementary aspects of user state and behavior~\cite{nicolas2012brain, gesture2006, icpr2000}. Leveraging these heterogeneous data sources enables richer and more adaptive interfaces, ultimately improving a system’s ability to understand, predict, and respond to human needs.

However, real-world multimodal data rarely exist in clean and controlled forms. Noise introduced by environmental factors, hardware constraints, and user variability often contaminates each modality~\cite{noise1, noise2, hand1990}. Label quality is inconsistent, especially in tasks involving subjective cognitive states, where human self-reporting can be unreliable~\cite{aBCI3, aBCI}. These challenges make it difficult for multimodal learning systems to remain stable and robust, particularly when deployed outside laboratory settings. As a result, building models that can effectively handle uncertainty—stemming from noisy data, imperfect labels, and modality heterogeneity—has become a central obstacle in advancing robust multimodal AI for BCI~\cite{cvpr2020}.

These issues become more complex when the heterogeneous nature of multimodal signals is considered. Each modality follows distinct statistical patterns and responds differently to environmental disturbances~\cite{body2007}. Some modalities may momentarily degrade in quality, while others remain reliable or lack the capacity to capture subtle semantic cues~\cite{aBCI2}. This modality heterogeneity often leads to inconsistencies between modalities, making the learning process highly sensitive to modality-specific noise. When noisy or unreliable modalities dominate training, they can distort the learned representation space, ultimately weakening the system’s reliability and interpretability~\cite{Li2025MultimodalBCI}.

To address these challenges, recent research has increasingly emphasized the importance of cross-modal semantic consistency~\cite{MultiModal1}. Despite their differences, multimodal signals that capture the same underlying event or human state should share a coherent semantic structure. By projecting heterogeneous modalities into a shared latent space, it becomes possible to reduce modality gaps and expose meaningful structural relationships across modalities~\cite{Multimodal2}. Such shared structures not only facilitate robust representation learning but also provide a foundation for identifying uncertainty: when modalities fail to align in the latent space, this misalignment may signal noise, ambiguity, or semantic inconsistency. Consequently, latent-space consistency offers a powerful mechanism for both enhancing representation quality and reasoning about the reliability of the data~\cite{crosskd2024, KDD2024}.

Nevertheless, existing approaches have limitations in operationalizing this idea. Knowledge distillation based methods transfer semantic knowledge from a more expressive modality or model to a weaker one, but they often struggle to handle uncertainty arising from inconsistent modalities~\cite{kd-hinton, kd-survey}. Active learning approaches aim to reduce annotation cost by selecting informative samples, but they typically rely on uncertainty estimated from a single modality or model output, overlooking cross-modal structure that could better inform sample acquisition~\cite{pr2024, nn2022}. These limitations highlight the need for a framework that not only aligns multimodal representations but also leverages their consistency to reason about uncertainty and data quality in a more principled manner.

Motivated by these observations, this thesis presents an integrated approach to uncertainty-resilient multimodal learning via consistency-guided cross-modal transfer. The core idea is straightforward: use the semantic consistency shared across modalities as a central learning signal. By designing a framework in which heterogeneous signals are aligned in a shared latent space, the system can extract robust semantic patterns, identify unreliable or noisy samples, and selectively enhance the learning process even under imperfect supervision~\cite{prototype2024, cross2012, select2007}.

Within this framework, the thesis investigates two complementary directions for leveraging cross-modal consistency. The first focuses on uncertainty-resilient representation learning, exploring how knowledge transfer from semantically richer modalities can stabilize representations under noisy or ambiguous supervision~\cite{CMKD2022, crosskd2024}. The second examines data-efficient multimodal learning, where cross-modal consistency is used to identify informative samples and mitigate the impact of unreliable annotations~\cite{AL2024}. Although these directions differ in emphasis, they share the central idea that aligning heterogeneous modalities in a shared latent structure provides a principled foundation for robust multimodal learning.

To examine the effectiveness and broader implications of this framework, the thesis evaluates its two complementary components across distinct learning perspectives. The first component focuses on uncertainty-resilient representation learning and investigates how semantic structure can be transferred from a more expressive modality to stabilize representations in noisy and heterogeneous multimodal conditions~\cite{CMKD2023ICLR}. The second component explores data-efficient multimodal learning by analyzing how cross-modal consistency can guide sample selection and reduce uncertainty in active learning scenarios~\cite{DL2021, video2001}. Together, these evaluations offer a comprehensive view of how consistency-guided cross-modal transfer can enhance both robustness and efficiency in multimodal systems, outlining its potential to support adaptive and reliable BCI and HCI applications.

\chapter{Related Work}

\section{Cross-Modal Knowledge Distillation}
Knowledge Distillation (KD)~\cite{kd-survey} is a widely used technique for transferring knowledge from a high-capacity teacher model to a lightweight student model, with the aim of improving the latter’s performance without significantly increasing its complexity. According to a recent practical survey on KD, existing approaches can be broadly categorized into logit-based, feature-based, and adaptive methods~\cite{CMKD2023ICLR}.

Logit-based KD methods~\cite{kd-hinton} distill knowledge by directly aligning the output logits of the student model with those of the teacher. This approach is the most straightforward and widely adopted, as it encourages the student to mimic the teacher’s predictive distribution. However, when the teacher and student operate on different modalities—as in cross-modal KD—direct logit matching alone may be insufficient for effective knowledge transfer, due to the modality gap and the lack of shared low-level representations.

Feature-based KD methods~\cite{ekd} address this issue by aligning intermediate representations between the teacher and student, often through transformation layers or projection modules. By matching features in a latent space, these methods can facilitate cross-modal knowledge transfer even when the input modalities differ, as the alignment occurs at a more abstract, modality-invariant level. In the context of EEG–vision KD, feature-based approaches are particularly relevant because they allow the EEG-based student to benefit from semantically rich visual representations while mitigating modality-specific discrepancies.

Adaptive KD methods~\cite{cmkd_ER} extend this concept by dynamically adjusting the distillation process based on the characteristics of each training sample. Rather than applying the same distillation strategy and weight uniformly across all samples, adaptive KD selectively emphasizes certain types of knowledge or modulates the distillation strength according to sample difficulty, teacher confidence, or other criteria. This is especially useful for emotion recognition, where data uncertainty and label noise can vary significantly between samples. By tailoring the distillation process, adaptive KD can prevent the student from overfitting to noisy or uninformative teacher signals, thereby improving robustness and generalization.

In this work, we incorporate elements of both feature-based and adaptive KD into a cross-modal framework that transfers knowledge from a vision-based teacher to an EEG-based student. Specifically, we align modality-invariant representations through a prototype-based similarity module (feature-based KD) while adapting the distillation strength using uncertainty scores derived from EEG differential entropy (DE) features (adaptive KD). This design allows our framework to address the modality gap and sample-level variability inherent in EEG emotion recognition.

\section{Multimodal Active Learning}
Active learning (AL)~\cite{DL2021} aims to reduce labeling costs by selectively querying the most informative samples, thereby enabling efficient data acquisition under limited annotation budgets. Traditional AL strategies primarily rely on uncertainty-based criteria—such as entropy, margin sampling, or Bayesian uncertainty—to determine which samples require additional labels. However, when applied to real-world affective or cognitive systems, these unimodal AL methods often suffer from noisy observations, ambiguous human states, and the inherent subjectivity of emotional responses. Such limitations hinder the reliability of uncertainty estimates and degrade the effectiveness of conventional querying strategies.

Multimodal, active learning has emerged as an alternative approach that leverages complementary information across multiple modalities to improve query selection and reduce uncertainty. By integrating diverse signals—such as EEG, facial expressions, speech, physiological responses, and user interactions—multimodal AL methods can better capture latent affective states and mitigate the ambiguity present in individual modalities~\cite{MultiModal1}. These approaches typically aim to exploit cross-modal consistency, where aligned information from one modality helps refine or verify inferences made from another. As a result, multimodal AL frameworks can more reliably identify samples that contain high informational value, ultimately improving the efficiency of the learning process~\cite{Multimodal2}.

In addition to passive multimodal fusion, recent works explore interactive or human-in-the-loop paradigms, where a model actively seeks clarification from users when cross-modal cues conflict or when its own uncertainty surpasses a threshold. This perspective treats user feedback as an additional modality that provides corrective information, enabling the system to refine affective or cognitive interpretations through iterative querying. Such interaction-driven AL designs are especially effective in domains where ground-truth labels are subjective, sparse, or difficult to collect, as they enable targeted refinement of uncertain or ambiguous instances.

Furthermore, several multimodal AL methods explicitly incorporate semantic or feature-level alignment to better guide sample selection. These approaches evaluate uncertainty not only at the prediction level but also at the representation level, allowing the system to detect inconsistencies across modalities and actively query samples whose multimodal embeddings consistently show high divergence. This strategy results in more robust sample selection, especially in tasks where one modality is significantly noisier than others.

Overall, multimodal active learning provides a powerful framework for reducing uncertainty in emotionally or cognitively rich environments. By combining cross-modal information, interactive querying, and uncertainty-aware strategies, it enables models to efficiently acquire labels and refine their interpretations of complex human states. This paradigm complements the knowledge distillation approaches discussed in Section 2.1, as both leverage cross-modal signals to correct uncertain or ambiguous representations—ultimately contributing toward the goal of uncertainty-resilient multimodal learning.

\chapter{Methods}

\section{Uncertainty-Aware Representation Learning via Knowledge Distillation}

In this paper, we address the problem of learning robust EEG representations under label noise and data uncertainty by distilling and transferring knowledge gained from the visual modality. The dataset contains paired EEG and visual samples $\{x_s^i, x_t^i, y^i\}_{i=1}^N$. Here, $x_s$ and $x_t$ represent data from the student (EEG) and teacher (visual) modalities, respectively, and $y$ denotes the corresponding ground truth labels. The feature extractors $\mathcal{E}$ and $\mathcal{V}$ map raw EEG and visual signals into their respective feature spaces. Also, EEG label noise lead to discrepancies between the ground truth labels $y$ and the distillation target from the teacher model, which negatively affects the student model's training.
Our goal is to learn a function $f_\theta : \mathcal{X}_s \rightarrow \mathcal{Y}$ that maps EEG representations to labels while leveraging visual knowledge. 
This is achieved by minimizing the total loss $\mathcal{L}$, which integrates all loss components in our framework.

\begin{sidewaysfigure*}
    \centering
    \includegraphics[width=\linewidth]{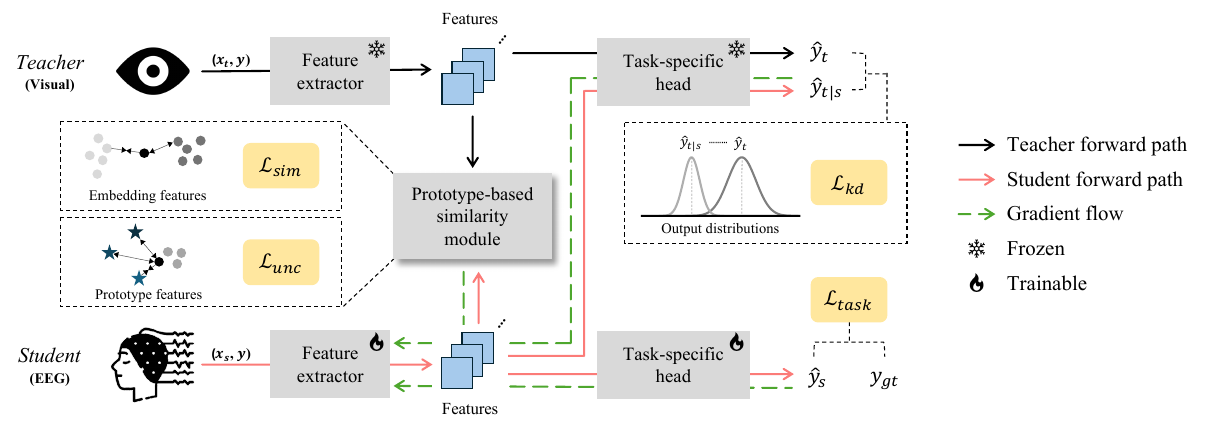}
    \caption{Overview of the proposed cross-modal KD framework. Intermediate features are extracted from both the teacher and student networks via their respective feature extractors. These features are passed through a prototype-based similarity module, which facilitates evidence-based uncertainty estimation and class-level semantic alignment. Additionally, the intermediate feature from the student is injected into the pre-trained teacher's task-specific distillation head to obtain predictions $\hat{y}_{t|s}$, which are then compared with the teacher's own predictions $\hat{y}_t$.}
    \label{fig1}
\end{sidewaysfigure*}

\subsection{Prototype-Based Similarity Module}

This work aims to mitigate modality discrepancy and label noise in EEG-based learning. To begin with, we introduce the prototype-based similarity module, which builds upon prior work~\cite{prototype2024} using prototype-based similarity to improve cross-modal feature alignment.
In our framework, this module performs prototype-based similarity alignment guided by Dirichlet-based uncertainty estimation, allowing the model to align modality-specific representations while explicitly accounting for semantic ambiguity. Given EEG samples $x_s$ and visual samples $x_t$, we extract their features $f_s$, $f_t$ and corresponding embeddings $\mathbf{e}_s$, $\mathbf{e}_t$ as follows:
\begin{equation}\label{eq:feat}
    f_s,\mathbf{e}_s = \mathcal{E}(x_s), \ f_t,\mathbf{e}_t = \mathcal{V}(x_t), 
\end{equation} 
where $\mathcal{E}$ and $\mathcal{V}$ denote the EEG and visual feature extractors, respectively. 
We construct a similarity matrix $Q$ using cosine similarity to measure the pairwise relationships between EEG and visual embeddings, which serves as the basis for cross-modal feature alignment and is computed as:
\begin{equation}\label{eq:cosim}
Q_{ii}(\mathbf{e}_s, \mathbf{e}_t)=\beta \cdot \left({\frac{\mathbf{e}_s^i}{||\mathbf{e}_s^i||}}\right) \cdot \left({\mathbf{e}_t^i \over ||\mathbf{e}_t^i||}\right)^T.
\end{equation}

Here, $Q\in\mathbb{R}^{N \times N}$ denotes the pairwise similarities between EEG and visual samples in a batch, and $\beta$ is a temperature parameter. The diagonal elements $Q_{ii}$ correspond to matching EEG-visual pairs, while the off-diagonal elements $Q_{ij}$ indicate non-matching pairs. To align EEG representations with visual features, we define the similarity-based loss function using a softmax-based contrastive (InfoNCE) formulation:
\begin{equation}\label{eq:loss_sim}
\mathcal{L}_{sim} = - {1 \over N} \sum_{i=1}^N \log {{\exp(Q_{ii}(\mathbf{e}_s, \mathbf{e}_t))} \over {\sum_{k=1}^N \exp(Q_{ik}(\mathbf{e}_s, \mathbf{e}_t))}}.
\end{equation}

By optimizing $\mathcal{L}_{sim}$,  the model is encouraged to align EEG and visual features by maximizing the similarity between matched pairs.  
However, this loss does not account for the semantic ambiguity arising from label noise and misalignment issues. To address this limitation, we incorporate Dirichlet-based uncertainty estimation, which quantifies the model's confidence in each sample by measuring its similarity to class prototypes $\Phi$. 
Specifically, the concentration parameter of the Dirichlet distribution is computed as $\alpha_i = \exp(Q(\mathbf{e}, \Phi) / \tau) + 1$ for each class $i$, where the similarity matrix is defined as:
\begin{equation}\label{eq:cosim2}
Q_{ij}(\mathbf{e}, \Phi) = \beta \cdot \left({\mathbf{e}^i \over ||\mathbf{e}^i||}\right) \cdot \left({\Phi^j \over ||\Phi^j||}\right)^T.
\end{equation}

This formulation estimates sample reliability via the alignment between embeddings and class prototypes $\Phi$, where the total evidence is computed as $e_i = \sum_{j=1}^c \alpha_{ij}$ with $c$ denoting the number of prototypes. The uncertainty is calculated from this evidence, with $\tau$ as a temperature parameter:
\begin{equation}\label{eq:unc}
u_i= 1-{c \over e_i}= 1-{c \over {\sum_{j=1}^c (\exp({Q_{ij} \over \tau})+1)}}. 
\end{equation}

As the total evidence increases, the corresponding aleatoric uncertainty decreases, indicating higher confidence in the prediction. Thus, $u$ reflects the inverse of total evidence and serves as a principled measure of aleatoric uncertainty.
To integrate this uncertainty into training, we define the uncertainty-aware loss that aligns model-inferred uncertainty with observed similarity-based uncertainty. In this loss, $h_j$ denotes the average similarity of sample $j$ to all non-matching samples, weighted by a scaling factor $\delta$:
\begin{equation}\label{eq:loss_unc}
\mathcal{L}_{unc} = {1 \over N} \sum_{j=1}^N || u_j- (\delta h_j)||^2_2.
\end{equation}

\subsection{Task-Specific Head}

Subsequently, we employ a task-specific distillation head. Inspired by~\cite{crosskd2024}, which reinterprets student features through the teacher’s structure in a unimodal setting, we extend this idea to a cross-modal scenario, where the teacher and student operate in different modalities. To minimize the semantic discrepancy caused by the modality gap, we feed the student’s intermediate features into the teacher’s head, specifically at the layer just before the final projection. This allows the student to benefit from the teacher’s role in guiding semantic alignment, even when the teacher’s predictions do not fully correspond to the ground truth. Accordingly, the outputs of the teacher and student heads are defined as follows:
\begin{equation}\label{eq:logits}
\hat{y}_t = \mathcal{H}_t(f_t), \ \hat{y}_s = \mathcal{H}_s(f_s), 
\end{equation}
where $\mathcal{H}_t$ and $\mathcal{H}_s$ denote the task-specific distillation heads of the teacher and student models, respectively. The extracted teacher feature $f_t$ is passed through all layers of $\mathcal{H}_t$ to produce its output. In contrast, the student feature $f_s$ is processed not only by the student's own head $\mathcal{H}_s$, but also injected into the teacher's head $\mathcal{H}_t$. Specifically, instead of entering from the first layer, $f_s$ is inserted at an intermediate fully connected layer $l$ and propagated through the remaining layers. This operation is defined as:
\begin{equation}\label{eq:logits2}
\hat{y}_{t|s} = \mathcal{H}_t^{(l:)}(f_s).
\end{equation}

Here, $\mathcal{H}_t^{(l:)}$ represents the teacher’s task-specific distillation head from layer $l$ onward. This design ensures that the student’s features are not directly forced to mimic the teacher’s low-level representations. Rather, they are aligned with higher-level decision-making processes that incorporate semantic information. To encourage the student's extracted feature $f_s$ to produce predictions similar to the teacher’s output, we apply the Kullback-Leibler divergence (KL) loss between the outputs $\hat{y}_t$ and $\hat{y}_{t|s}$. This is referred to as the distillation loss and is defined as:
\begin{equation}\label{eq:loss_kd}
\mathcal{L}_{kd} = KL (\hat{y}_t||\hat{y}_{t|s}) = \sum \hat{y}_t \log {\hat{y}_t\over \hat{y}_{t|s}}.
\end{equation}

This objective promotes the convergence of the student model toward the teacher’s semantic representations, even across heterogeneous modalities. Although $\mathcal{L}_{kd}$ facilitates this semantic alignment, additional mechanisms are required to adapt the model to the specific task. To ensure that the model learns task-specific information, we apply appropriate objective functions for regression and classification tasks, respectively. For the DEC task, we adopt the standard cross-entropy (CE) loss between the student’s predictions $\hat{y}_s$ and the ground truth labels $y$. This explicit supervision helps the student model follow the teacher’s guidance toward the correct label space, particularly under noisy or uncertain conditions. The CE loss is defined as:
\begin{equation}\label{eq:loss_ce}
\mathcal{L}_{task} = CE(\hat{y}_s, y) = - \sum y \log{\hat{y}_s}.
\end{equation}

In contrast, the CER task is handled using the concordance correlation coefficient (CCC) loss~\cite{v2e} to directly measure the agreement between the predicted continuous values $\hat{y}_s$ and the ground truth $y$. This loss effectively encourages the model to match both the mean and the variance of the predictions to the target distribution. The CCC loss is given by:
\begin{equation}
\label{eq:loss_ccc}
\mathcal{L}_{task} 
= CCC(\hat{y}_s, y) 
= {2\sigma_{{\hat{y}_s}y} 
\over \sigma^2_{\hat{y}_s} + 
\sigma^2_{y} + 
(\mu_{\hat{y}_s}-\mu_y)^2 },
\end{equation}
where $\sigma_{{\hat{y}_s}y}$ denotes the corresponding covariance, $\sigma_{\hat{y}_s}$ and $\sigma_y$ the variances, and $\mu_{\hat{y}_s}$ and $\mu_y$ the means, respectively.

\subsection{Overall Objective}

The overall loss function for our proposed method is defined as follows:
\begin{equation}\label{eq:loss_total}
\mathcal{L} = \lambda_1 \mathcal{L}_{sim} + 
\lambda_2 \mathcal{L}_{unc} + 
\lambda_3 \mathcal{L}_{kd} + 
\lambda_4 \mathcal{L}_{task},
\end{equation}
where $\lambda_1$, $\lambda_2$, $\lambda_3$, and $\lambda_4$ are hyperparameters that balance the contributions of similarity preservation, uncertainty regularization, knowledge distillation, and task-specific supervision, respectively.
An overview of the overall procedure and model architecture is illustrated in Fig.~\ref{fig1}.

\section{Cross-Modal Consistency-Guided Active Learning}

\begin{figure}[t!]
\centerline{\includegraphics[scale=0.8]{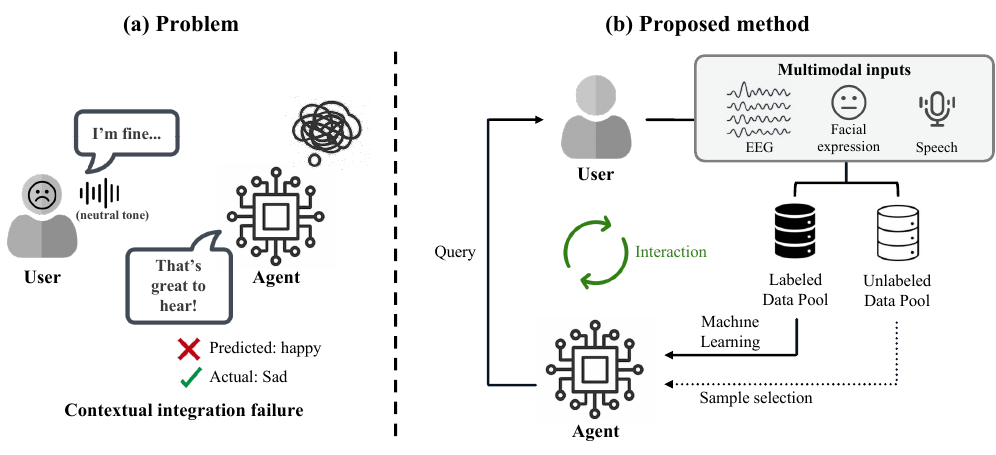}}
\caption{
Overview of the proposed cross-modal AL framework for enhancing emotional intelligence in multimodal human–agent interaction.
(a) \textbf{Problem:} The agent with limited affective understanding misinterprets the user’s emotional states from multimodal signals, resulting in inaccurate emotion recognition and poor adaptability.
(b) \textbf{Proposed method:} The proposed framework enables the agent to improve emotional understanding via uncertainty-based active learning that selectively queries the user for feedback and updates its model.
}
\label{fig_ovrall_al}
\end{figure}

The proposed framework jointly integrates a multimodal consistency module and uncertainty-driven active querying into a human-in-the-loop learning paradigm.
As illustrated in Fig. ~\ref{fig_ovrall_al}, this framework iteratively refines affective understanding by training on labeled data, estimating uncertainty on unlabeled samples, and querying the most informative instances for annotation to progressively expand the labeled pool.
During training, it leverages multimodal signals to capture rich affective context, while at deployment, the system operates solely with EEG input, while maintaining practical deployability.

\subsection{Multimodal Consistency Module}
The system receives synchronized multimodal inputs $x = \{ x_{\text{EEG}}, x_{\text{Face}} \}$, and the goal is to infer the user's affective state $y$. The dataset is partitioned into a labeled pool $D_L = \{(x_i, y_i)\}$ and an unlabeled pool $D_U = \{x_j\}$. Training begins by optimizing the model on $D_L$. To build a coherent emotional representation across heterogeneous signals, we introduce a multimodal consistency module that aligns EEG and facial embeddings within a shared latent space.
Each modality $m \in \{\text{EEG}, \text{Face}\}$ is processed by a modality- specific encoder $E_m(\cdot)$, producing a latent representation,
\begin{equation}
    z_m = E_m(x_m).
\end{equation}

To encourage consistent emotional representations across the two modalities, we initially project the latent embeddings into a shared normalized space,
\begin{equation}
    \hat{z}_m = \frac{z_m}{\| z_m \|_2},
\end{equation}
and compute a pairwise cross-modal similarity matrix,
\begin{equation}
    S(i,j) = \hat{z}_{\text{EEG}}^{(i)\top} \hat{z}_{\text{Face}}^{(j)}.
\end{equation}

The diagonal elements measure the agreement between synchronized EEG–face pairs, while the off-diagonal elements capture mismatched or redundant correlations. We enforce alignment through a symmetric contrastive loss:
\begin{equation}
    \mathcal{L}_{\text{sim}} 
    = \frac{1}{2} \Big(
        \mathrm{CE}(S, y)
        + 
        \mathrm{CE}(S^\top, y)
    \Big).
\end{equation}

While $\mathcal{L}_{\text{sim}}$ aligns paired EEG--face embeddings at the batch level, 
it treats all samples as equally reliable. 
However, affective signals frequently contain ambiguous or weakly expressive cues,
and certain samples exhibit poor cross-modal agreement even after alignment.
To account for this variability, we derive from the similarity matrix a 
sample-wise \emph{cross-modal reliability target} that quantifies how well the 
EEG embedding of each sample agrees with facial embeddings in the same batch.
For the $i$-th EEG embedding, we measure its average similarity to all facial
embeddings except its own paired sample:
\begin{equation}
    h_i = \frac{1}{N-1} \sum_{j \neq i} S(i,j).
\end{equation}

A large value of $h_i$ indicates that the EEG embedding resembles many unrelated facial embeddings. Conversely, a smaller $h_i$ implies that the sample exhibits modality-specific characteristics and is more reliable for alignment. To ensure scale invariance across batches, we normalize the scores using min--max normalization: 
\begin{equation}
    \tilde{h}_i = 
    \frac{h_i - \min(h)}{\max(h) - \min(h) + \varepsilon}.
\end{equation}

We then define the target reliability as follows, providing a supervisory signal:
\begin{equation}
    r_i^{\ast} = 1 - \tilde{h}_i,
\end{equation}
assigning lower reliability to samples exhibiting large cross-modal disagreement.
The model produces modality-specific reliability estimates 
$r^{\text{EEG}}_i$ and $r^{\text{Face}}_i$, which are trained to match the supervisory reliability signal:
\begin{equation}
    \mathcal{L}_{\text{rel}}
    = \frac{1}{2} \left(
        \| r^{\text{EEG}} - r^{\ast} \|_2^2
        +
        \| r^{\text{Face}} - r^{\ast} \|_2^2
    \right).
\end{equation}

While the multimodal consistency module effectively enforces cross-modal feature alignment, the EEG head performs affective state estimation based solely on EEG embeddings, which are available at deployment. 
Given an EEG embedding $z^{EEG}$, the EEG-specific head $H(\cdot)$ predicts the affective state $\hat{y}=H(z^{EEG})$. 
For classification-based emotion recognition, the task supervision is applied through the cross-entropy loss:
\begin{equation}
    \mathcal{L}_{task} = - \sum_{c=1}^{C} y_c \log \hat{y}_c,
\end{equation}
where $C$ denotes the number of classes and $y_c$ is the target.

\subsection{Active Query Strategy}
Once the model has been optimized on the labeled pool $\mathcal{D}_L$, it is used to estimate uncertainty on the remaining unlabeled samples $\mathcal{D}_U$. The objective of the active query strategy is to identify the most informative instances whose annotation is expected to yield maximal performance improvement. Importantly, the uncertainty used for querying is derived solely from the predictive distribution of the classifier trained on multimodally aligned features.

Given an unlabeled sample $x_j \in \mathcal{D}_U$, the model produces its class probability distribution $\hat{p}_j = f(x_j)$. We quantify the uncertainty of the prediction using the Shannon entropy of the output distribution:
\begin{equation}
    \mathcal{U}(x_j)
    = - \sum_{c=1}^{C} 
        \hat{p}_j^{(c)} \log \hat{p}_j^{(c)}.
\end{equation}

Higher entropy indicates lower confidence and thus greater potential value for annotation. At each active learning iteration, all unlabeled samples are ranked by their uncertainty values. Let $\tau$ denote the acquisition ratio.  
We select the most uncertain instances by taking the top-$\tau$ fraction of the ranked set:
\begin{equation}
    \mathcal{Q}
    = \mathrm{Top}\text{-}\tau
      \big\{
        x_j \in \mathcal{D}_U \,\mid\, \mathcal{U}(x_j)
      \big\}.
\end{equation}

The queried samples $\mathcal{Q}$ are sent to the human oracle for annotation and transferred from $\mathcal{D}_U$ to the labeled pool:
\begin{align}
    \mathcal{D}_L &\leftarrow
    \mathcal{D}_L \cup \{(x_j, y_j)\}_{x_j \in \mathcal{Q}}, \\
    \mathcal{D}_U &\leftarrow
    \mathcal{D}_U \setminus \mathcal{Q}.
\end{align}

This iterative process enables the model to concentrate labeling effort on the most informative regions of the data space, resulting in a more sample-efficient gradual improvement of affective understanding.

\subsection{Overall Objective and Training Loop}
The overall framework jointly optimizes affective representation alignment, cross-modal reliability modeling, and task-specific prediction. The total training objective integrates the three components as:
\begin{equation}
    \mathcal{L}_{\text{total}}
    =
    \lambda_{1}\, \mathcal{L}_{\text{sim}}
    +
    \lambda_{2}\, \mathcal{L}_{\text{rel}}
    +
    \lambda_{3}\, \mathcal{L}_{\text{task}},
\end{equation}
where $\lambda_{1}$, $\lambda_{2}$, and $\lambda_{3}$ act as weighting factors that regulate the relative contributions of similarity alignment, reliability regularization, and affective classification.

During each active learning iteration, the model is first optimized on the current labeled pool $\mathcal{D}_L$ using $\mathcal{L}_{\text{total}}$. The updated model then estimates uncertainty over the unlabeled pool $\mathcal{D}_U$, and the top-$p$ \% most uncertain samples are queried from the human oracle. The newly annotated instances are added to $\mathcal{D}_L$ while removed from $\mathcal{D}_U$, enabling the model to refine its affective understanding.

\chapter{Experiments}

\section{Experimental Settings}
\sloppy
\paragraph{Datasets.} MAHNOB-HCI~\cite{mahnob} is a widely used multimodal dataset for emotion recognition and implicit tagging research based on behavioral and physiological responses. It includes recordings from 30 participants who watched affective video clips lasting 34 to 117 seconds. During each session, EEG signals were recorded using 32 electrodes, and behavior was captured with 6 cameras. Subjects 12, 15, and 26 did not complete data collection, so only 27 participants are used in this study. Each trial includes self-reported ratings from 1 to 9 for arousal, valence, dominance, and predictability, and valence was additionally annotated by experts to capture temporal emotional dynamics. Therefore, the dataset enables research on both emotion classification and regression tasks.

\paragraph{Task.} Emotion recognition tasks are generally categorized into two main approaches: discrete emotion classification (DEC) and continuous emotion regression (CER)~\cite{aBCI}. DEC involves categorizing emotional states into predefined classes such as happiness, sadness, anger, or fear. In contrast, CER represents emotions as points in a continuous space, typically defined by the valence–arousal model. While DEC is often easier to interpret and evaluate in terms of categorical emotion labels, CER enables the modeling of subtle and dynamic emotional changes for more precise emotional analysis.

\paragraph{Implementation Details.} The overall architecture employs an EEG-Conformer~\cite{eegconformer} and a CNN-based transformer as the feature extractors for the student and teacher models, respectively.
The model was trained using the Adam optimizer with the learning rate ranging from $10^{-3}$ to $10^{-6}$ for 100 epochs. The dataset was partitioned using 5-fold cross-validation with a group-by-trial strategy. 
To improve generalization, early stopping with a patience of 20 epochs was applied during training in each fold. The entire framework was implemented in PyTorch and trained on NVIDIA GPUs to accelerate computation.

\paragraph{Evaluation Metrics.}
We compare our method with baseline approaches on both discrete and continuous emotion recognition tasks. For the DEC task, we evaluate performance using accuracy and macro F1 score, while for the CER task, we report root mean squared error (RMSE), Pearson correlation coefficient (PCC), and concordance correlation coefficient (CCC) as evaluation metrics. We then present feature space visualizations to demonstrate the effectiveness of our approach, followed by an ablation study analyzing the contribution of each loss component of our framework.

\begin{table*}[!t]
    \centering
    \caption{Performance comparison of our proposed method with baseline unimodal and multimodal methods for discrete emotion classification (DEC).}
    \label{table:dec}
    \renewcommand{\arraystretch}{1.1}
    \begin{threeparttable}
    \resizebox{\textwidth}{!}{
    \begin{tabular}{c c c c c c}
        \toprule[1.3pt]
        \multicolumn{2}{c}{\multirow{2}{*}[-0.5em]{\centering Method}}
        & \multicolumn{2}{c}{Arousal}
        & \multicolumn{2}{c}{Valence} \\
        \cmidrule(lr){3-4} \cmidrule(lr){5-6}
        & & Acc. $\pm$ std. & F1. $\pm$ std. & Acc. $\pm$ std. & F1. $\pm$ std. \\
        \midrule[1.1pt]

        \multirow{4}{*}{\centering Unimodal}
         & DeepConvNet\cite{deepconvnet}    & 44.0$\pm$1.52 & 42.1$\pm$1.77 & 45.7$\pm$2.36 & 44.2$\pm$1.76 \\
         & EEGNet\cite{eegnet}              & 43.6$\pm$1.87 & 47.4$\pm$1.46 & 43.1$\pm$2.01 & 49.2$\pm$2.94 \\
         & EEGFormer\cite{eegformer}        & 45.2$\pm$3.49 & 40.8$\pm$3.24 & 48.1$\pm$5.08 & 50.2$\pm$4.82 \\
         & MASA-TCN~\cite{masa}             & 46.4$\pm$1.32 & 48.7$\pm$1.56 & 45.8$\pm$2.67 & 50.9$\pm$4.65 \\
        \midrule

        \multirow{4}{*}{\centering Multimodal}
         & CAFNet~\cite{caf}                & \textbf{59.4$\pm$1.17} & 59.6$\pm$1.83 & 55.6$\pm$3.02 & 57.2$\pm$2.59 \\
         & Visual-to-EEG~\cite{v2e}         & 53.9$\pm$3.34 & 54.6$\pm$2.71 & 50.3$\pm$1.42 & 54.6$\pm$1.12 \\
         & EmotionKD~\cite{ekd}             & 56.8$\pm$2.57 & 52.2$\pm$3.14 & 49.9$\pm$2.24 & 52.0$\pm$2.28 \\
         & \textbf{Ours}                    & 57.1$\pm$2.38 & \textbf{60.0$\pm$1.45} & \textbf{57.9$\pm$3.21} & \textbf{59.4$\pm$2.47} \\
        \bottomrule[1.3pt]
    \end{tabular}}
    \begin{tablenotes}
        \item Acc.: accuracy ($\%$), F1.: F1 score ($\%$).
    \end{tablenotes}
    \end{threeparttable}
\end{table*}


\section{Classification Performance}

To validate the effectiveness of our proposed method, we compare it with both unimodal and multimodal baselines on the MAHNOB-HCI dataset. Table~\ref{table:dec} summarizes the classification performance for the DEC task and the regression performance for the CER task. Here, we report the DEC results, focusing on accuracy and F1 scores for arousal and valence, where our method shows consistent improvements over unimodal baselines. We compare against DeepConvNet~\cite{deepconvnet}, EEGNet~\cite{eegnet}, EEGFormer~\cite{eegformer}, and MASA-TCN~\cite{masa}, including models designed for EEG-based emotion recognition and those commonly used in general EEG modeling, among which MASA-TCN demonstrates competitive performance. However, our model achieves superior results, with 57.1$\%$ accuracy and the highest F1 score of 60$\%$ in arousal classification.
Beyond unimodal models, our method continues to demonstrate strong performance compared to multimodal baselines, such as CAFNet~\cite{caf}, Visual-to-EEG~\cite{v2e}, and EmotionKD~\cite{ekd}, which incorporate additional modalities for affective state decoding from EEG signals.
Although CAFNet reports higher accuracy in arousal classification, our model records the highest F1 score for arousal and outperforms all baselines in both accuracy and F1 score for valence classification. This comprehensive superiority indicates that our method achieves the most balanced and reliable performance across emotion categories.

\section{Regression Performance}

\begin{table*}[t!]
    \centering
    \caption{Performance comparison of our proposed method with baseline unimodal and multimodal methods for continuous emotion regression (CER).}
    \label{table:cer}
    \renewcommand{\arraystretch}{1.1}
    \begin{threeparttable}
    \resizebox{\textwidth}{!}{
    \begin{tabular}{c c c c c}
        \toprule[1.3pt]
        \multicolumn{2}{c}{\multirow{2}{*}[-0.5em]{\centering Method}}
        & \multicolumn{3}{c}{Valence} \\
        \cmidrule(lr){3-5}
        & & RMSE $\pm$ std. & PCC $\pm$ std. & CCC $\pm$ std. \\
        \midrule[1.1pt]

        \multirow{4}{*}{\centering Unimodal}
         & DeepConvNet\cite{deepconvnet} & 0.049$\pm$0.030 & 0.323$\pm$0.045 & 0.255$\pm$0.034 \\
         & EEGNet\cite{eegnet}           & 0.047$\pm$0.035 & 0.316$\pm$0.078 & 0.301$\pm$0.079 \\
         & EEGFormer\cite{eegformer}     & 0.050$\pm$0.029 & 0.334$\pm$0.056 & 0.238$\pm$0.072 \\
         & MASA-TCN~\cite{masa}          & 0.058$\pm$0.035 & 0.390$\pm$0.044 & 0.338$\pm$0.045 \\
        \midrule
        \multirow{4}{*}{\centering Multimodal}
         & CAFNet~\cite{caf}             & 0.046$\pm$0.062 & 0.368$\pm$0.058 & 0.309$\pm$0.060 \\
         & Visual-to-EEG~\cite{v2e}      & 0.055$\pm$0.028 & \textbf{0.462$\pm$0.052} & \textbf{0.372$\pm$0.084} \\
         & EmotionKD~\cite{ekd}          & 0.052$\pm$0.034 & 0.422$\pm$0.085 & 0.310$\pm$0.056 \\
         & \textbf{Ours}                 & \textbf{0.043$\pm$0.027} & 0.449$\pm$0.075 & 0.359$\pm$0.069 \\
        \bottomrule[1.3pt]
    \end{tabular}
    }
    \begin{tablenotes}
        \item RMSE: root mean squared error, PCC: Pearson correlation coefficient, \\CCC: concordance correlation coefficient.
    \end{tablenotes}
    \end{threeparttable}
\end{table*}
We evaluate CER performance using RMSE, PCC, and CCC metrics for valence prediction, as summarized in Table~\ref{table:cer} based on the MAHNOB-HCI dataset. Lower RMSE indicates better prediction accuracy, while higher PCC reflects stronger correlation with the ground truth. Among all compared models, our proposed method achieves the lowest RMSE of 0.043, and the second-highest PCC and CCC scores of 0.449 and 0.359, respectively.
Compared to unimodal baselines, our model shows clear improvements across all three metrics. Among unimodal methods, MASA-TCN~\cite{masa} demonstrates competitive performance but still falls short of our approach. In the multimodal setting, Visual-to-EEG~\cite{v2e} demonstrates strong correlation metrics, although its performance is less stable in terms of prediction error. In contrast, our model attains superior balance, achieving both the lowest RMSE and strong correlation metrics.
Taken together, the results demonstrate that our approach effectively captures both discrete and continuous emotional representations. We attribute this gain to extracting semantically aligned features and making robust predictions under noisy EEG conditions.

\section{Feature Space Visualization}

\begin{figure}[H]
\centerline{\includegraphics[scale=0.5]{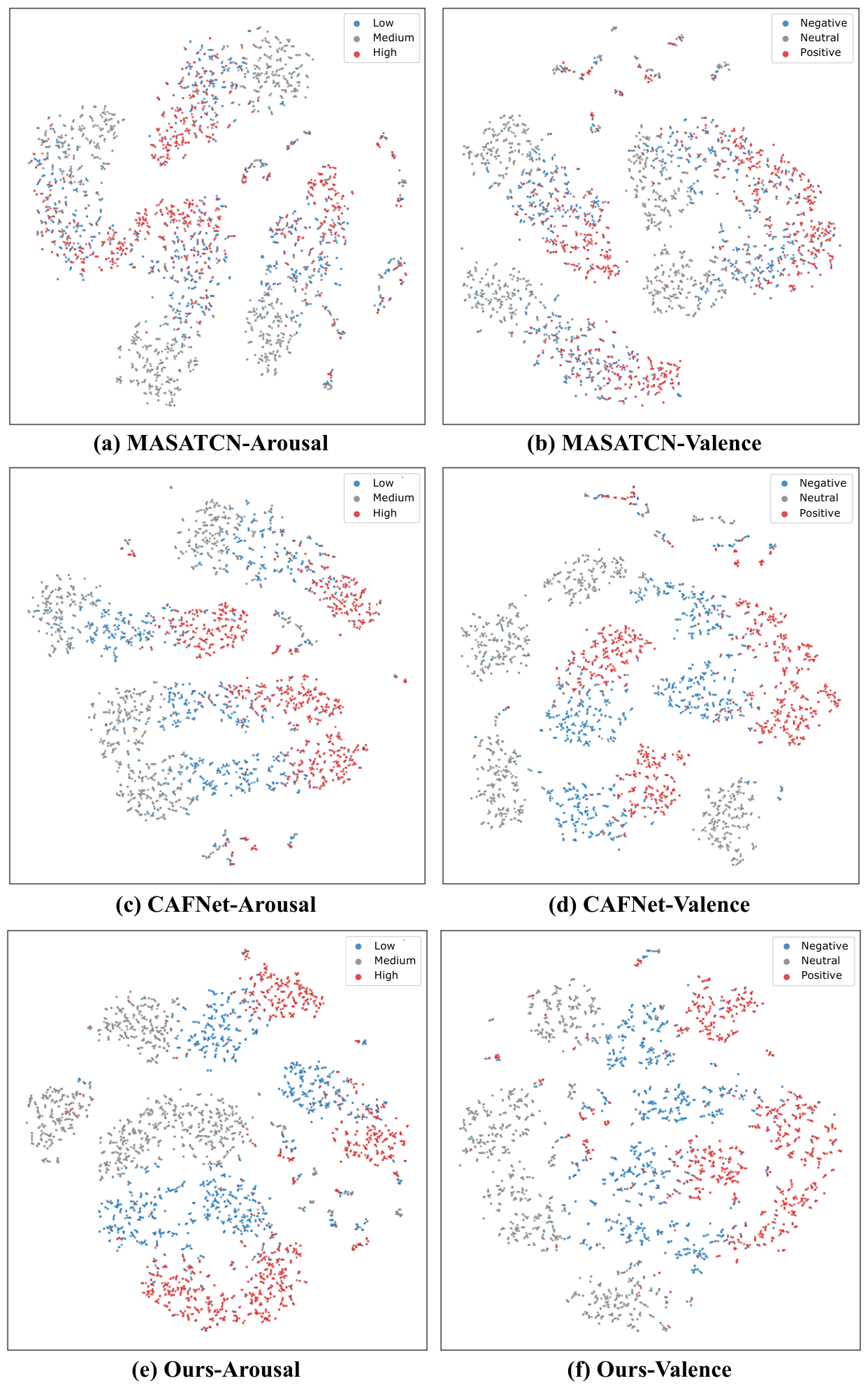}}
\caption{Visualization of the learned feature distribution. For comparison, we present the features learned by the unimodal sota model MASA-TCN (top), the multimodal sota CAFNet (middle), and our proposed cross-modal KD method (bottom) for both the arousal and valence classification tasks.}
\label{fig:tsne_comparison}
\end{figure}

In order to verify the effectiveness of our framework, we visualize the learned feature distributions projected onto a two-dimensional space, as shown in Fig.~\ref{fig:tsne_comparison}.
We compare our method against a strong unimodal baseline and a competitive multimodal model, focusing on feature representations learned during arousal and valence classification tasks.
We first examine the projected feature spaces learned by MASA-TCN in Fig. 2(a) and 2(b), which reveal scattered and overlapping clusters. This indicates weak separability between low and high emotional states, suggesting limited discriminative power in the learned EEG representations. Fig. 2(c) and 2(d) show the feature spaces learned by CAFNet, demonstrating improved clustering compared to MASA-TCN, particularly for arousal. However, class overlap still persists highlighting challenges in modeling subtle emotional variations. In contrast, Fig. 2(e) and 2(f) illustrate that our method yields well-separated and compact clusters in both emotional dimensions. In particular, the valence feature distributions demonstrate clear separation among negative, neutral, and positive emotional states.  
Moreover, they exhibit tight class grouping, which reflects stronger consistency and enhanced discriminative capability. These visual patterns support the quantitative results reported in Fig.~\ref{fig:tsne_comparison}, highlighting that our method consistently outperforms MASA-TCN and CAFNet across most evaluation metrics.

\begin{table}[t]
    \centering
    \caption{Ablation study of the proposed loss components on the overall performance for the DEC task.}
    \label{table:ablation}
    \renewcommand{\arraystretch}{1}
    \begin{threeparttable}
    \begin{tabular}{c c c c c c c}
        \toprule[1.3pt]
        \multirow{2}{*}{$\mathcal{L}_{sim}$} & 
        \multirow{2}{*}{$\mathcal{L}_{unc}$} & 
        \multirow{2}{*}{$\mathcal{L}_{kd}$} & 
        \multicolumn{2}{c}{Arousal} & 
        \multicolumn{2}{c}{Valence} \\
        \cmidrule(lr){4-5} \cmidrule(lr){6-7}
        & & & Acc. $\pm$ std. & F1. $\pm$ std. & Acc. $\pm$ std. & F1. $\pm$ std. \\ 
        \midrule[1.1pt]
        \checkmark &  &                       & 42.8$\pm$4.94 & 44.2$\pm$7.45 & 41.6$\pm$4.36 & 41.9$\pm$4.73 \\
        & \checkmark &                        & 43.0$\pm$1.80 & 43.2$\pm$2.12 & 42.8$\pm$2.97 & 42.9$\pm$3.32 \\
        &  & \checkmark                       & 45.3$\pm$2.63 & 47.3$\pm$2.88 & 44.4$\pm$7.91 & 43.6$\pm$2.11 \\
        \checkmark & \checkmark &             & 51.0$\pm$3.61 & 53.7$\pm$0.94 & 43.9$\pm$6.95 & 44.1$\pm$7.24 \\
        \checkmark &  & \checkmark            & 54.6$\pm$3.97 & 53.2$\pm$2.26 & 48.4$\pm$4.51 & 48.5$\pm$4.66 \\
        & \checkmark & \checkmark             & \textbf{58.6}$\pm$1.79 & 58.9$\pm$1.28 & 51.4$\pm$8.23 & 57.3$\pm$2.79 \\
        \checkmark & \checkmark & \checkmark  & 57.1$\pm$2.38 & \textbf{60.0$\pm$1.45} & \textbf{57.9$\pm$3.21} & \textbf{59.4$\pm$2.47} \\
        \bottomrule[1.3pt]       
    \end{tabular}
    \end{threeparttable}
\end{table}

\begin{table}[!t]
    \centering
    \caption{Ablation study of the proposed loss components on the overall performance for the CER task.}
    \label{table:ablation2}
    \renewcommand{\arraystretch}{1}
    \begin{threeparttable}
    \begin{tabular}{c c c c c c}  
        \toprule[1.3pt]
        \multirow{2}{*}{$\mathcal{L}_{sim}$} & \multirow{2}{*}{$\mathcal{L}_{unc}$} & \multirow{2}{*}{$\mathcal{L}_{kd}$} & \multicolumn{3}{c}{Valence} \\
        \cmidrule(lr){4-6}
        & & & RMSE $\pm$ std. & PCC $\pm$ std. & CCC $\pm$ std. \\ 
        \midrule[1.1pt]
        \checkmark &  &                          & 0.057$\pm$0.018 & 0.251$\pm$0.064 & 0.226$\pm$0.052 \\
        & \checkmark &                           & 0.046$\pm$0.012 & 0.300$\pm$0.068 & 0.266$\pm$0.082  \\
        &  & \checkmark                          & 0.053$\pm$0.016 & 0.267$\pm$0.037 & 0.243$\pm$0.077 \\
        \checkmark & \checkmark &                & 0.052$\pm$0.020 & 0.315$\pm$0.057 & 0.305$\pm$0.058 \\
        \checkmark &  & \checkmark               & 0.050$\pm$0.012 & 0.357$\pm$0.066 & 0.343$\pm$0.072\\
        & \checkmark & \checkmark                & 0.049$\pm$0.017 & 0.406$\pm$0.080  & \textbf{0.369$\pm$0.078}\\
        \checkmark & \checkmark & \checkmark     & \textbf{0.043$\pm$0.027} & \textbf{0.449$\pm$0.075} & 0.359$\pm$0.069  \\
        \bottomrule[1.3pt]      
    \end{tabular}
    \end{threeparttable}
\end{table}

\section{Performance Comparison Across Loss Configurations}

To examine the contribution of each loss component, we conducted an ablation study, as summarized in Table~\ref{table:ablation} for the DEC task and Table~\ref{table:ablation2} for the CER task. We evaluated three loss terms in our framework: similarity-based loss $\mathcal{L}_{sim}$, uncertainty-aware loss $\mathcal{L}_{unc}$, and knowledge distillation loss $\mathcal{L}_{kd}$.
In the DEC task, $\mathcal{L}_{kd}$ demonstrates the greatest individual impact on classification performance. When used alone, it outperforms both $\mathcal{L}_{sim}$ and $\mathcal{L}_{unc}$, underscoring its effectiveness in guiding the student model via teacher supervision.
Meanwhile, $\mathcal{L}_{unc}$ improves robustness to ambiguous EEG signals and shows strong synergy with $\mathcal{L}_{kd}$, with their combination yielding the highest arousal accuracy. However, the best overall performance is achieved when all three components are used together.
In the CER task, $\mathcal{L}_{unc}$ exhibits the most significant impact applied in isolation.
Among the two-loss combinations, the pair of $\mathcal{L}_{sim}$ and $\mathcal{L}_{kd}$ delivers the strongest performance, indicating effective interaction between feature similarity and teacher supervision. Still, this configuration falls short compared to the full combination of all three losses.
Across both DEC and CER tasks, $\mathcal{L}_{kd}$ consistently proves to be the most influential component. This trend underscores its central role in enabling effective cross-modal supervision. Incorporating all three objectives leads to more robust and generalizable emotion recognition, especially under ambiguous EEG conditions.

\section{Label Efficiency Across Training Configurations}

\begin{table}[t]
\centering
\caption{Comparison of label efficiency under different training configurations.}
\begin{tabular}{lcccccc}
\toprule
\textbf{Method} & \textbf{10 \%} & \textbf{30 \%} & \textbf{50 \%} 
& \textbf{70 \%} & \textbf{100 \%} \\
\midrule
No-AL & -- & -- & -- & -- & 0.597 \\
Random & 0.462 & 0.476 & 0.526 & 0.669 & 0.812 \\
Ours (Uncertainty) & 0.462 & 0.460 & \textbf{0.584} & 0.716 & 0.853 \\
\bottomrule
\end{tabular}
\label{table:label_efficiency}
\end{table}

To evaluate the label-efficiency benefits of the proposed framework, we compare it against two alternative training configurations, as summarized in Table~\ref{table:label_efficiency} \textbf{No-AL} denotes a full-supervision baseline that uses 100 \% of the labeled data from the start, without multimodal consistency learning or active querying. \textbf{Random} enables the multimodal consistency module but acquires new samples uniformly at random. \textbf{Ours} represents the complete system, combining cross-modal consistency learning with uncertainty-driven sample acquisition. Across all labeling budgets, our method achieves higher accuracy while requiring substantially fewer labeled samples. Notably, at only 50 \% of the labels, our method surpasses both the Random baseline and even the fully supervised No-AL model, demonstrating the effectiveness of jointly leveraging multimodal alignment and uncertainty-aware query selection.

\section{Effect of Uncertainty Reduction During Active Learning}

\begin{figure}[t!]
\centerline{\includegraphics[scale=0.5]{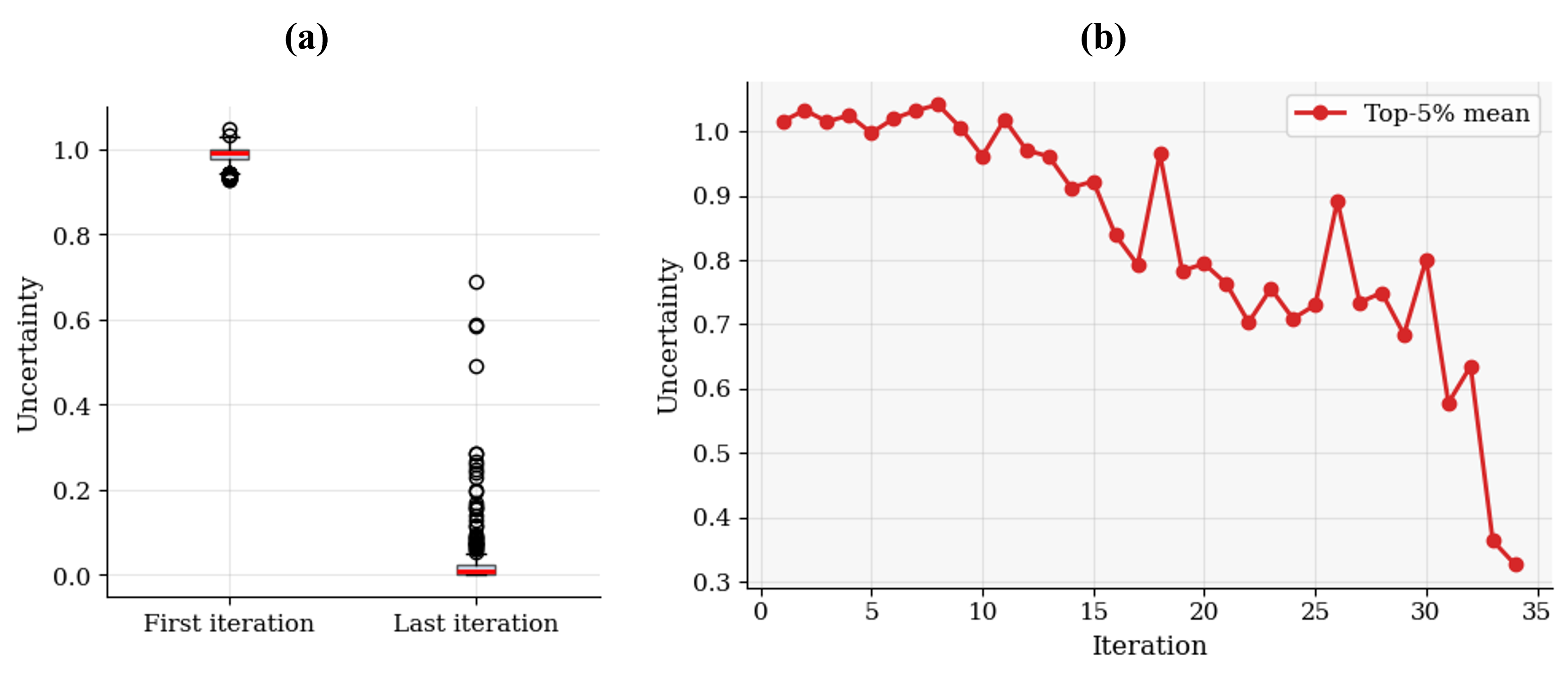}}
\caption{Evolution of uncertainty during cross-modal active learning.
(a) Comparison of uncertainty distributions between the first and last iterations.
(b) Trend of the top-5 \% uncertainty mean across active learning steps.}
\label{fig:fig_unc_score}
\end{figure}

To verify that our method effectively reduces model uncertainty over time, we examine how the predicted uncertainty scores evolve throughout the active learning loop.
Fig.~\ref{fig:fig_unc_score}(a) compares the uncertainty distribution of unlabeled samples at the beginning and the end of active learning. The first iteration exhibits high and widely spread uncertainty values, whereas the final iteration shows a sharply compressed distribution near zero, indicating that the remaining unlabeled samples have become substantially easier for the model.
Fig.~\ref{fig:fig_unc_score}(b) tracks the mean uncertainty of the
top--5 \% most uncertain samples selected at each iteration. The steady downward trend demonstrates that the queried samples become progressively less uncertain, confirming that the proposed strategy successfully reduces overall uncertainty within the pool.
In summary, this experiment validates that our uncertainty-based acquisition mechanism effectively drives down model uncertainty across iterations, ensuring that labeling efforts consistently target the most informative samples.

\chapter{Conclusion}

This thesis presented a unified approach to addressing the challenges of uncertainty in multimodal learning, particularly within domains such as brain–computer interfaces and human–computer interaction. Real-world multimodal data are frequently affected by noise, imperfect labels, and heterogeneous sensing conditions, all of which hinder the stability and generalization of conventional learning systems. In response, this work proposed consistency-guided cross-modal transfer as an overarching framework for constructing multimodal models that remain robust and reliable under imperfect supervision.
The framework was explored along two complementary directions. The first focused on uncertainty-resilient representation learning, examining how semantic structure from a richer modality can be transferred to a weaker one to stabilize representations in noisy or ambiguous settings. The second direction investigated data-efficient multimodal learning, leveraging cross-modal consistency as a signal to guide selective annotation and reduce uncertainty within active learning scenarios. Although these two approaches target different aspects of multimodal learning, they are connected through the shared principle that semantic consistency across modalities can serve as a reliable foundation for both robust representation learning and uncertainty-aware supervision.
Through experimental analyses, this thesis demonstrated that consistency-guided multimodal learning can improve representational stability, highlight informative samples, and better capture the underlying semantic structure shared across modalities. These findings extend beyond simple fusion strategies and provide evidence that modality relationships themselves can be used to interpret uncertainty—a capability of particular value in real-world BCI and HCI applications.
In summary, this thesis provides an integrated perspective on building uncertainty-resilient multimodal systems by combining semantic consistency, knowledge transfer, and data-efficient learning. The proposed framework highlights how cross-modal relationships can be leveraged not only to enhance robustness but also to reason about the reliability of data itself, offering a meaningful step toward more adaptive, trustworthy, and practical multimodal intelligence for future BCI and HCI technologies.

\end{document}